\documentclass[runningheads]{llncs}
\pdfoutput=1
\usepackage[T1]{fontenc}
\usepackage{graphicx}
\usepackage{amsmath}
\usepackage{amssymb}
\usepackage{booktabs}
\usepackage{array}
\usepackage{tabularx}
\usepackage{multirow}
\usepackage{url}
\usepackage{xcolor}
\usepackage[hidelinks]{hyperref}
\usepackage{enumitem}

\let\oldthebibliography\thebibliography
\renewcommand{\thebibliography}[1]{\oldthebibliography{#1}\setlength{\itemsep}{0pt}\setlength{\parskip}{0pt}}
\providecommand{\doi}[1]{\url{https://doi.org/#1}}
\begin{document}
\title{A Dual-Stream Regulated Reconstruction and Segmentation Network with Hierarchical Artifact-Prior Modeling for Ultra-Low-Field Pediatric Neuroimaging}
\titlerunning{Dual-Stream Regulated Network for Low-Field Pediatric MRI}
\author{Bahram Jafrasteh\inst{1} \and Leo Milecki\inst{1} \and Qingyu Zhao\inst{1}}
\authorrunning{B. Jafrasteh et al.}
\institute{Department of Radiology, Weill Cornell Medicine, New York, NY, USA \\
\email{baj4003@med.cornell.edu}}
\maketitle
\begin{abstract}
Automated quality assessment, enhancement, and segmentation of multiple structures in $0.064\,\mathrm{T}$ ultra-low-field pediatric MRI are limited by a low signal-to-noise ratio, weak anatomical boundaries, and frequent artifacts. We present a unified framework for the LISA 2026 Challenge that performs all three tasks together within one inference pipeline. A network with two coupled streams, built on a 3D U-Net, first reconstructs an enhanced uLF volume and then combines the original and enhanced images for subcortical segmentation. To improve boundary stability, we add an auxiliary class covering brain tissue outside the target structures, derived from whole brain masks. A head conditioned on an artifact graph predicts the seven artifact ratings from reconstruction residuals and frozen segmentation features. We address the scarcity of dense annotations using diffeomorphic registration from atlas to target for label propagation and to regularize anatomical reconstruction. We report validation results across all three tasks.

\end{abstract}
\section{Introduction}

During the first years of life, the brain undergoes rapid morphological and tissue changes. Accurate magnetic resonance imaging (MRI) is therefore essential for characterizing healthy neurodevelopment and enabling the early identification of neurodevelopmental disorders~\cite{tapp2025qa}. While quality assessment and segmentation methods perform well on adult MRI, they often fail on pediatric scans because of reduced contrast between gray and white matter and rapidly changing anatomy~\cite{tapp2025qa,zalevskyi2025segmenting}. These challenges are further amplified at low field strength. Portable ultra-low-field (uLF) systems such as the $0.064\,\mathrm{T}$ Hyperfine SWOOP scanner improve accessibility in low- and middle-income settings~\cite{deoni2021accessible,sheth2021assessment}, but their lower spatial resolution, reduced signal-to-noise ratio (SNR), and frequent acquisition artifacts make quality assessment and anatomical segmentation considerably more difficult~\cite{tapp2025qa,zalevskyi2025segmenting}.

The LISA 2026 Challenge addresses these limitations through three complementary tasks on pediatric uLF MRI acquired across three international sites: \textbf{Task~1a} predicts the severity of seven artifact types (Noise, Zipper, Positioning, Banding, Motion, Contrast, and Distortion); \textbf{Task~1b} enhances uLF images by reducing noise and motion artifacts; and \textbf{Task~2} segments eleven clinically relevant subcortical structures.

Although these tasks are inherently related, existing challenge solutions treat them independently using separate models~\cite{tapp2025qa,zalevskyi2025segmenting}. We instead propose a \emph{unified} framework that jointly performs quality assessment, image enhancement, and anatomical segmentation through a shared representation while still allowing each task to be optimized on its own terms. Our contributions are:
\begin{enumerate}[label=\roman*)]
    \item A 3D U-Net with two coupled streams, linking reconstruction and segmentation through an operator that isolates gradients between them, stabilizing joint optimization while sharing parameters.
    \item A semi-supervised strategy driven by registration that fine-tunes a pretrained diffeomorphic network to propagate atlas labels onto unlabeled scans while regularizing anatomical consistency during reconstruction.
    \item A head for quality prediction, conditioned on a graph, exploiting correlations among the seven artifact attributes through an ordinal loss that accounts for distance between severity levels, operating on detached embeddings from residuals and skip connections.
    \item An auxiliary class covering brain foreground outside the target structures, which regularizes subcortical segmentation by separating target structures, remaining brain tissue, and extracranial background.
    \item A training strategy with three stages: a reconstruction warm-up, joint training of reconstruction and segmentation with registration-driven augmentation, and a final quality-head optimization, yielding one deployable model for all tasks.
\end{enumerate}

\section{Methods}

\begin{figure}[t]
\centering
\includegraphics[width=\textwidth]{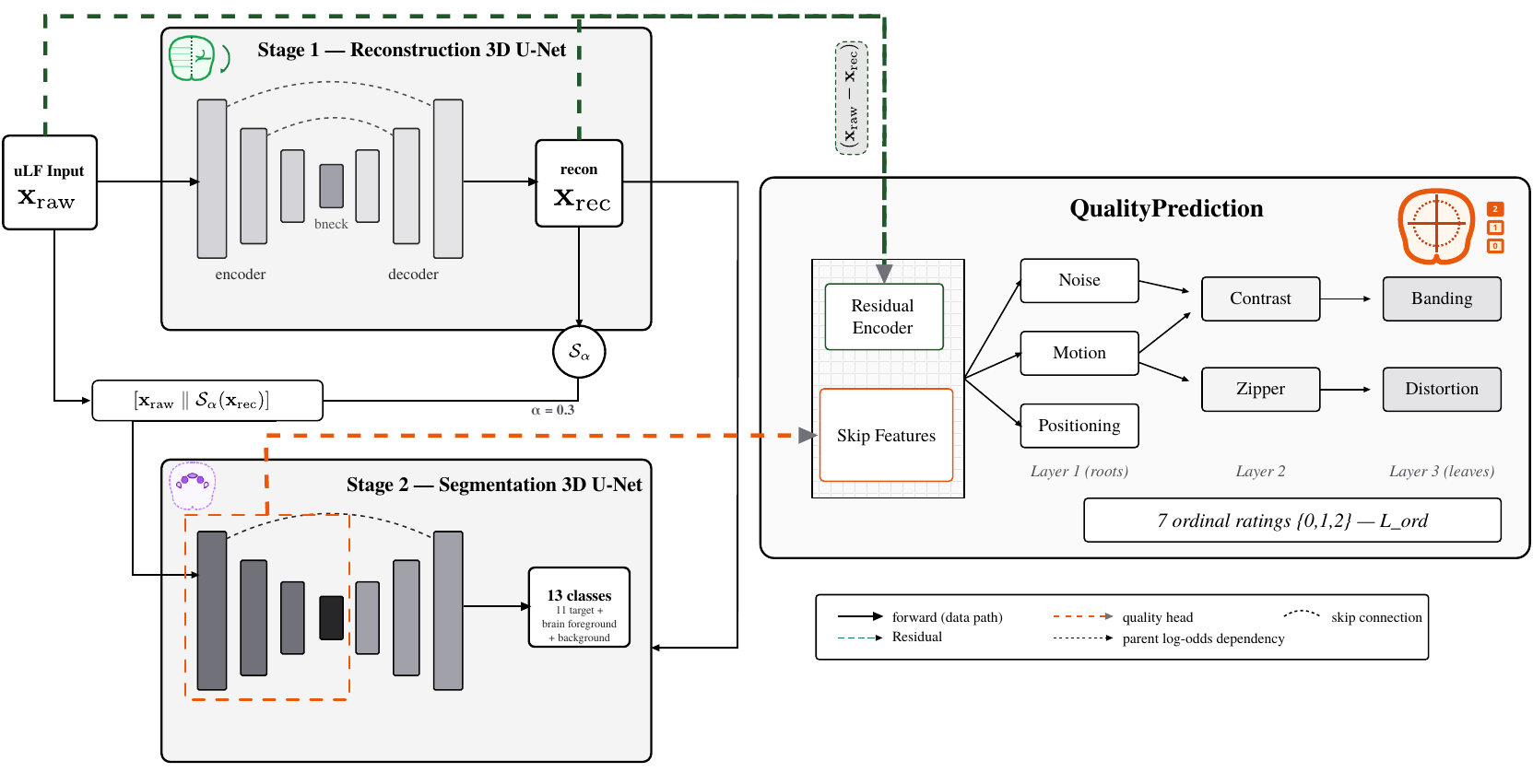}
\caption{Overview of the proposed framework. Stage~1 reconstructs an enhanced volume from the degraded uLF input; the enhanced volume is concatenated with the raw input (through the operator $\mathcal{S}_\alpha$ that scales gradients, $\alpha{=}0.3$) and fed to the Stage~2 segmentation encoder-decoder. In a final stage, the hierarchical quality head is trained on embeddings from two sources:
a residual encoder processes the artifact residual map
$(\mathbf{x}_{\mathrm{raw}} - \mathbf{x}_{\mathrm{rec}})$, and the detached
features from the segmentation encoder's skip connections provide anatomy-aware
context (gradient stopped, dashed).}
\label{fig:arch}
\end{figure}

\textbf{Overview.}
Our framework jointly addresses image enhancement (reconstruction, Task~1b), anatomical segmentation (Task~2), and quality assessment (Task~1a) within one unified architecture (Figure~\ref{fig:arch}). The framework consists of a shared backbone for reconstruction and segmentation, followed by a lightweight quality assessment head. For stability, training is performed sequentially, where the joint reconstruction and segmentation backbone is first trained (Section~\ref{sec:joint}). Then, the quality assessment head is trained on the frozen backbone features. The segmentation head predicts 13 mutually exclusive classes: the eleven official LISA structures, an auxiliary class for brain foreground outside those structures, and background. The auxiliary class contains only voxels from the brain mask that lie outside the target structures. To enlarge the pool of densely annotated anatomy, we additionally incorporate high-field neonatal scans from the developing Human Connectome Project (dHCP)~\cite{makropoulos2018developing}. Fully labeled LISA and dHCP samples supervise segmentation directly, while unlabeled uLF scans serve as registration targets to generate warped pairs of atlas images and labels that augment segmentation training.

Throughout the paper, we use \emph{degradation} to denote the seven LISA artifact types: Noise, Zipper, Positioning, Banding, Motion, Contrast, and Distortion. Each image is associated with a vector of artifact severities $\mathbf{m}_i\in\{0,1,2\}^{7}$, where entries denote absent, mild, or severe artifacts. For real LISA scans, this vector is derived from expert quality ratings when available. To augment the data for training quality assessment, we also synthetically degraded scans by a degradation pipeline (Section~\ref{sec:data}).

\subsection{Joint Reconstruction--Segmentation}
\label{sec:joint}

Enhancement and segmentation are complementary objectives but require different optimization goals. We therefore use a 3D U-Net with two coupled streams and controlled gradient flow.
The backbone is a pair of 3D U-Nets~\cite{cicek20163d} operating on the same input size.

\subsubsection{Reconstruction.}

Reconstruction aims not only to denoise uLF MRI, but to suppress artifact patterns while preserving anatomical structure. Because several uLF artifacts have strong signatures in the frequency domain, we combine spatial, spectral, structural, and smoothness constraints.
Stage~1 (reconstruction) maps the single-channel uLF input $\mathbf{x}_{\mathrm{raw}}$ to an enhanced volume $\mathbf{x}_{\mathrm{rec}}$ using isotropic $3{\times}3{\times}3$ convolutions, four levels of down-sampling, and dedicated adapters on the skip connections that decouple encoder features from the decoder path. 

Losses computed pixel by pixel are insufficient for ultra-low-field MRI because several artifacts (e.g., banding and zipper artifacts) exhibit characteristic patterns in the frequency domain that are only weakly penalized by spatial losses. We therefore define a reconstruction loss combining a spatial MSE with a term in the frequency domain that matches magnitude spectra from the Fourier transform. Let $\mathcal{F}$ denote the orthonormal 3D Fourier transform, $\mathcal{L}_{\mathrm{spatial}}=\lVert\mathbf{x}_{\mathrm{rec}}-\mathbf{x}_{\mathrm{clean}}\rVert_2^2$ and $\mathcal{L}_{\mathrm{freq}}=\big\lVert|\mathcal{F}(\mathbf{x}_{\mathrm{rec}})|-|\mathcal{F}(\mathbf{x}_{\mathrm{clean}})|\big\rVert_2^2$.
From the metadata on artifact severity, we define an indicator of periodic artifacts that comes either from expert ratings for real LISA scans or from the simulator log for synthetically degraded scans. For synthetically degraded scans, the degradation pipeline records the applied corruption types and relative magnitudes; these metadata are used only for reconstruction training and loss weighting, and are not treated as Task~1a true quality labels. The reconstruction fidelity term is then

\begin{equation}
\mathcal{L}_{\mathrm{fidelity}} = (1 - w_f)\,\mathcal{L}_{\mathrm{spatial}} + w_f\,\mathcal{L}_{\mathrm{freq}}
\label{eq:recmse}
\end{equation}
The frequency weight $w_f$ is set from the metadata on artifact severity: $w_f=0.4$ when banding or zipper artifacts are present, and $w_f=0.15$ otherwise. This gives stronger spectral supervision to artifacts with periodic structure in the frequency domain, while keeping the spatial term dominant for other samples. This adaptive fidelity term is combined with structural and smoothness regularizers,

\begin{equation}
\mathcal{L}_{\mathrm{rec}} = 1.5\,\mathcal{L}_{\mathrm{fidelity}} + 1.5\,\mathcal{L}_{\mathrm{SSIM}} + 0.5\,\mathcal{L}_{\mathrm{TV}},
\end{equation}
where $\mathcal{L}_{\mathrm{SSIM}}$~\cite{wang2004ssim} preserves local structure and $\mathcal{L}_{\mathrm{TV}}$ suppresses residual noise at high frequencies. Reconstruction supervision comes from paired clean and degraded data, including synthetically degraded volumes whose clean targets are known by construction.

\subsubsection{Segmentation.}
Stage~2 receives a two-channel volume formed by concatenating the raw input with the Stage~1 output, so that the segmentation encoder can use both the original signal and its enhanced counterpart.
In uLF pediatric MRI, poor contrast between gray and white matter makes subcortical boundaries difficult to delineate. We therefore introduce an auxiliary class for brain foreground to provide coarse anatomical context and to discourage target structures from leaking into the extracranial background.
Stage~2 predicts the eleven LISA target structures, i.e., left/right hippocampus, ventricle, caudate, lentiform, and thalamus, plus the corpus callosum. To stabilize boundaries under poor contrast between gray and white matter, we add an auxiliary class for brain foreground outside the target structures and train the decoder with $n_{\mathrm{classes}}{=}13$: eleven target structures, non-target brain foreground, and background. The auxiliary class is derived from the whole brain mask after removing the eleven target structures. Target voxels retain their anatomical labels, remaining voxels inside the brain are assigned to the auxiliary class, and voxels outside the brain are assigned to the background. This yields a mutually exclusive label map with 13 classes that can be trained with standard cross-entropy and Dice losses over multiple classes. This design is consistent with recent evidence that subdividing heterogeneous background regions can improve biomedical segmentation by providing more informative contextual supervision~\cite{saluja2026backsplit}.

Rather than annotating this foreground prior manually, we obtain the whole brain mask by applying the pretrained MGA-Net neonatal model for brain extraction~\cite{jafrasteh2024mga}, designed for MRI data of low quality, to each uLF volume. All generated masks were visually inspected; fewer than 20 cases showed minor errors, which were manually corrected using MELAGE~\cite{jafrasteh2023melage,jafrasteh2024mga}. The resulting auxiliary class serves as a coarse spatial anchor, discouraging subcortical labels from leaking into the background outside the brain.
The segmentation objective combines weighted cross-entropy, a soft Dice loss over multiple classes, and a smoothness penalty based on total variation:
\begin{equation}
\label{eq:lossseg}
\mathcal{L}_{\mathrm{seg}} = \mathcal{L}_{\mathrm{CE}} + \mathcal{L}_{\mathrm{Dice}} + 0.03\,\mathcal{L}_{\mathrm{TV}} .
\end{equation}

\subsubsection{Registration-Driven Learning.}
\label{sec:reg}

Dense manual labels are limited in the LISA training set. We therefore use registration to propagate atlas labels to unlabeled uLF scans to enlarge the training pool for segmentation. In addition, since image enhancement should not modify the underlying anatomy, registration is used to regularize enhancement by penalizing anatomically implausible reconstructions.

We use a pretrained diffeomorphic registration network, in the style of VoxelMorph~\cite{balakrishnan2019voxelmorph}, for two complementary purposes.
First, it is fine-tuned during joint training to propagate labels from atlas to target: for uLF scans without dense subcortical labels, a labeled atlas $(\mathbf{x}_s,\mathbf{y}_s)$ matched in orientation is registered to the Stage~1 reconstruction $\mathbf{x}_{\mathrm{rec}}$, producing a warped pair of image and label $(\mathcal{T}_{\phi}(\mathbf{x}_s),\mathcal{T}_{\phi}(\mathbf{y}_s))$ which is added to the segmentation training batch as an augmented supervised sample. 

Second, a registration branch with gradients stopped serves as an evaluator of anatomical consistency for the reconstruction stream: for samples with a clean reference, large deformations required to align $\mathbf{x}_{\mathrm{rec}}$ to $\mathbf{x}_{\mathrm{clean}}$ indicate structural distortion and are penalized using
\begin{equation}
\mathcal{L}_{\mathrm{anat}} = \mathcal{L}_{\mathrm{NCC}}\!\big(\mathcal{T}_\phi(\mathbf{x}_{\mathrm{rec}}),\,\mathbf{x}_{\mathrm{clean}}\big) + \,\lVert \nabla \phi \rVert_2^2 .
\end{equation}
where $\mathcal{L}_{\mathrm{NCC}}$ denotes the normalized cross-correlation loss, and $\phi$ is the deformation field estimated by the frozen registration branch. The atlas pool includes labels from the LISA training set and 100 compatible dHCP samples. For both sources, we retain the eleven LISA target structures and the auxiliary mask for brain foreground outside the target structures.

\subsubsection{Overall Objective.}

To stabilize training on unlabeled or weakly labeled samples, we maintain a copy of the network, updated as an exponential moving average (EMA) of the trainable weights, that serves as a mean teacher~\cite{tarvainen2017meanteacher} providing consistency targets for the student. During joint training of reconstruction and segmentation, the main network minimizes a weighted sum of the per-task terms,
\begin{equation}
\mathcal{L}_{\mathrm{joint}} = \mathcal{L}_{\mathrm{seg}} + \mathcal{L}_{\mathrm{cons}} + \mathcal{L}_{\mathrm{rec}} + 0.2\,\mathcal{L}_{\mathrm{anat}},
\end{equation}
where $\mathcal{L}_{\mathrm{cons}}$ is the mean squared error between the student's and the EMA teacher's segmentation probabilities.

A key challenge when optimizing for multiple tasks is preventing the segmentation objective from interfering with the reconstruction network. We therefore insert an operator $\mathcal{S}_\alpha$ that scales gradients on the reconstructed branch before concatenation. $\mathcal{S}_\alpha$ is an identity map on the forward pass and multiplies incoming gradients by a constant $\alpha$ on the backward pass:
\begin{equation}
\mathbf{x}_{\mathrm{seg}} = \big[\,\mathbf{x}_{\mathrm{raw}} \;\|\; \mathcal{S}_\alpha(\mathbf{x}_{\mathrm{rec}})\,\big],
\qquad
\frac{\partial \mathcal{S}_\alpha(\mathbf{x})}{\partial \mathbf{x}} = \alpha\,\mathbf{I},\quad \alpha = 0.3 .
\end{equation}
This attenuates the segmentation gradient flowing back into the reconstruction stream, letting Stage~1 specialize in artifact suppression while still providing a useful enhanced input to Stage~2. The quality head is not coupled by gradients at all: it is trained in a separate final phase using detached embeddings, from the residual and from skip features, taken from the frozen backbone for reconstruction and segmentation (Section~\ref{sec:quality}).

\subsection{Quality Head}
\label{sec:quality}

Quality assessment differs from enhancement and segmentation because it requires explicit recognition of artifact type and severity. We therefore train a dedicated quality prediction head after freezing the backbone for reconstruction and segmentation, preventing the quality objective from perturbing the shared image representation.
The quality head predicts the seven ordinal artifact ratings using a dependency graph learned from statistics on how artifacts co-occur in the training set. Noise, Motion, and Positioning are treated as root attributes; Contrast and Zipper are conditioned on the root predictions; Banding and Distortion are conditioned on their respective parent nodes in the estimated dependency graph. Parent logits are detached before being passed to child predictors, so the child losses do not alter the parent classifiers.

The head uses two detached inputs: an embedding of the residual $\mathbf{x}_{\mathrm{raw}}-\mathbf{x}_{\mathrm{rec}}$ and an embedding that captures anatomical context, obtained by globally pooling skip features from the frozen segmentation encoder. These embeddings are concatenated with an embedding of acquisition orientation and passed through small MLP classifiers arranged according to the artifact graph.
Because severities are ordered ($0{<}1{<}2$), we use an ordinal loss that accounts for distance between severity levels. For artifact $a$ with logits $\mathbf{l}^{a}$, target $y^{a}$, and probabilities $p_j^a=\mathrm{softmax}(\mathbf{l}^{a})_j$,
\begin{equation}
\mathcal{L}_{\mathrm{ord}} =
\frac{1}{7}\sum_{a=1}^{7}\beta_a
\mathcal{L}_{\mathrm{CE}}(\mathbf{l}^{a},y^{a})
\left(1+\left|\sum_{j=0}^{2}j\,p_j^a-y^a\right|\right),
\label{eq:ord}
\end{equation}
where $\beta_a$ are weights for each artifact, normalized to sum to seven. These weights are used as a term that balances the tasks.

\section{Experimental Setup}

\subsection{Data, Preprocessing, and Validation}
\label{sec:data}

\textbf{Challenge data.}
The LISA challenge provides more than $500$ pediatric T2-weighted scans acquired on $0.064\,\mathrm{T}$ Hyperfine SWOOP scanners across three sites, with expert quality ratings for the seven Task~1 artifact domains and subcortical segmentations registered to uLF space for Task~2~\cite{tapp2025qa,lepore2025low}. All LISA uLF volumes were processed in the geometry specific to their acquisition. To avoid introducing unrealistic spatial assumptions, external scans were not treated as native isotropic inputs; instead, they were converted into orientation-specific geometries that resemble uLF acquisitions. External data were used only during training and were excluded from validation and test evaluation.

\textbf{External high-field data.}
To increase anatomical supervision, we incorporated 100 compatible neonatal MRI samples from the developing Human Connectome Project (dHCP). Only labels corresponding to the eleven LISA target structures and the class for non-target brain foreground were retained; all other tissue labels were discarded. To reduce the domain gap, dHCP images and labels were resampled into geometries compatible with LISA and synthetically degraded online during training.

\textbf{Synthetic degradation.}
Synthetic degradation was used to expose the model to a broader range of artifacts resembling uLF while preserving known clean targets. Degradation was applied online to the dHCP variants specific to each orientation and, with probability 0.5, to clean LISA scans, defined as uLF scans with expert QC ratings of 0 for all seven artifact attributes. The degradation pipeline stochastically composed physics corruptions resembling uLF, periodic artifacts, and effects on resolution and intensity, while recording the corresponding synthetic severity vector. Thus, dHCP contributed anatomical structure and clean references at high field strength, whereas the appearance of uLF and artifact severity were generated dynamically during training. Because synthetic corruption levels are not calibrated to the LISA ordinal QA scale, they were used only for reconstruction training; Task~1a was supervised only by expert ratings.

\textbf{Validation protocol.}
All validation splits were defined at the patient level to avoid leakage across original scans, synthetic degradations, variants specific to each orientation, and samples derived from atlases. Because more expert-rated samples were available for Task~1a, quality assessment used a validation split of 20\%, whereas reconstruction and segmentation used validation splits of 10\%, also defined at the patient level. Splits were stratified by labels specific to each task when available, with Task~1a stratified by the profile across all seven artifacts.

\subsection{Training Schedule}
\textbf{Overview.}
Training proceeded in three phases designed to stabilize the dependencies between tasks: reconstruction was first initialized independently, then reconstruction and segmentation were jointly optimized, and the quality head was finally trained on frozen backbone features.

\textbf{Phase~1: reconstruction warm-up.}
The network received degraded inputs and was trained to recover the corresponding clean targets using $\mathcal{L}_{\mathrm{rec}}$, with $\mathcal{L}_{\mathrm{anat}}$ added when a clean reference was available. The segmentation stage and quality head were not trained in this phase. This warm-up stabilized the enhanced image $\mathbf{x}_{\mathrm{rec}}$ before it was used as an input to the segmentation network.

\textbf{Phase~2: joint training of reconstruction and segmentation.}
Stage~1 and Stage~2 were optimized jointly using $\mathcal{L}_{\mathrm{joint}}$. The segmentation network received an input with two channels,
$[\mathbf{x}_{\mathrm{raw}}\parallel\mathcal{S}_{\alpha}(\mathbf{x}_{\mathrm{rec}})]$, where $\mathcal{S}_{\alpha}$ attenuated segmentation gradients flowing back into Stage~1. Fully labeled LISA and dHCP samples supervised $\mathcal{L}_{\mathrm{seg}}$ directly. For uLF samples without dense labels, the registration network aligned a labeled atlas matched in orientation to the target and added the resulting warped pair of atlas image and label to the segmentation batch as an augmented supervised sample. In parallel, we maintained a copy of the trainable network for reconstruction and segmentation, updated as an exponential moving average, that served as the mean teacher~\cite{tarvainen2017meanteacher}. During training, this teacher provided consistency targets for unlabeled or weakly labeled samples. After training, the EMA teacher weights were used as the final model for reconstruction and segmentation at inference.

\textbf{Phase~3: training the quality head.}
After Phase~2, the EMA teacher was retained as the final backbone for reconstruction and segmentation, and frozen. The hierarchical quality head was then trained on LISA samples using only expert artifact ratings. Its input was a detached embedding from two sources: the residual artifact representation from $\mathbf{x}_{\mathrm{raw}}-\mathbf{x}_{\mathrm{rec}}$ and pooled skip features from the frozen segmentation encoder. This separation prevented the quality objective from altering reconstruction or segmentation features and avoided the degenerate case, observed under joint training, in which every quality prediction collapsed to ``absent''.

\textbf{Implementation details.}
All networks were trained with Adam using an initial learning rate of \(10^{-3}\) and a polynomial schedule for decaying the learning rate, \(\eta_t=\eta_0(1-t/T)^{0.9}\), where \(T=50{,}000\) iterations. Training used a batch size of 4, and the EMA teacher used a decay of 0.999. Volumes were processed as full \(160\times192\times128\) inputs after preprocessing, without additional patch sampling.

\subsection{Evaluation Metrics}
Evaluation followed the LISA challenge protocol~\cite{tapp2025qa,maier2020bias}. Task~1a was assessed using accuracy, precision, recall, and F1/F2 scores for artifact severity prediction. Task~1b was evaluated using PSNR, FID~\cite{heusel2017gans}, LPIPS~\cite{zhang2018unreasonable}, BRISQUE$_\Delta$, CLIPIQA, and FRD. Task~2 was evaluated using Dice similarity coefficient (DSC), Hausdorff distance (HD), 95th-percentile Hausdorff distance (HD95), average symmetric surface distance (ASSD), and relative volume error (RVE).
\begin{table}[t]
\centering
\scriptsize
\setlength{\tabcolsep}{3pt}
\caption{Internal validation results. Segmentation: 7 volumes; enhancement: 12 volumes; quality assessment: 106 volumes. Distances in mm; RVE is absolute relative volume error.}
\label{tab:results}
\begin{tabularx}{\textwidth}{l*{5}{>{\centering\arraybackslash}X}}
\toprule
\multicolumn{6}{c}{\textbf{Task 2: Segmentation}} \\
Structure & DSC & HD & HD95 & ASSD & RVE \\
\midrule
Hippocampus L/R & .818/.830 & 2.51/2.79 & 1.70/1.46 & .37/.35 & .085/.056 \\
Ventricle L/R   & .858/.872 & 2.27/2.50 & 1.14/1.10 & .36/.34 & .044/.102 \\
Caudate L/R     & .887/.910 & 2.24/2.00 & 1.06/1.00 & .33/.28 & .071/.052 \\
Lentiform L/R   & .911/.919 & 1.88/1.70 & 1.14/1.06 & .36/.35 & .071/.063 \\
Thalamus L/R    & .935/.934 & 1.78/1.53 & 1.06/1.06 & .34/.34 & .024/.045 \\
Corpus callosum & .817 & 2.87 & 1.14 & .34 & .079 \\
\midrule
Mean            & .881 & 2.19 & 1.18 & .34 & .063 \\
\midrule
\multicolumn{6}{c}{\textbf{Task 1b: Enhancement}} \\
\multicolumn{6}{c}{PSNR: 34.17 \quad LPIPS: .160 \quad FID: $2.6{\times}10^{-5}$ \quad BRISQUE$_\Delta$: 4.23 \quad CLIPIQA: .257 \quad FRD: .174} \\
\midrule
\multicolumn{6}{c}{\textbf{Task 1a: Quality Assessment}} \\
 & Accuracy & Precision & Recall & F1 & F2 \\
\midrule
Weighted    & .829 & .682 & .633 & .638 & .632 \\
\midrule
Noise       & .877 & .739 & .862 & .784 & .825 \\
Zipper      & .811 & .773 & .748 & .760 & .752 \\
Positioning & .877 & .713 & .487 & .541 & .502 \\
Banding     & .972 & .660 & .663 & .662 & .663 \\
Motion      & .774 & .645 & .551 & .561 & .552 \\
Contrast    & .783 & .748 & .635 & .669 & .646 \\
Distortion  & .708 & .494 & .483 & .487 & .484 \\
\bottomrule
\end{tabularx}
\end{table}

\begin{table}[t]
\centering
\tiny
\setlength{\tabcolsep}{2.5pt}
\caption{Compact ablation summary (7 segmentation / 12 enhancement volumes). Distances in mm.}
\label{tab:ablation}
\begin{tabular}{lcccccccccc}
\toprule
Model       & DSC$\uparrow$ & HD95$\downarrow$ & ASSD$\downarrow$ & RVE$\downarrow$ & PSNR$\uparrow$ & LPIPS$\downarrow$ & FID$\downarrow$ & BRISQUE$_\Delta\downarrow$ & CLIPIQA$\uparrow$ & FRD$\downarrow$ \\
\midrule
Seg./Recon. only & .763 & 1.91 & .74 & .136 & 32.11 & .183 & $5.2{\times}10^{-5}$ & \textbf{1.71} & .226 & .201 \\
Detached         & .824 & 1.47 & .54 & .123 & 33.17 & .164 & $1.1{\times}10^{-4}$ & 3.23 & .242 & .193 \\
Proposed         & \textbf{.881} & \textbf{1.18} & \textbf{.34} & \textbf{.063} & \textbf{34.17} & \textbf{.160} & $\mathbf{2.6{\times}10^{-5}}$ & 4.23 & \textbf{.257} & \textbf{.174} \\
\bottomrule
\end{tabular}
\end{table}
\section{Results}

\begin{figure}[htbp]
\centering
\includegraphics[width=0.9\textwidth]{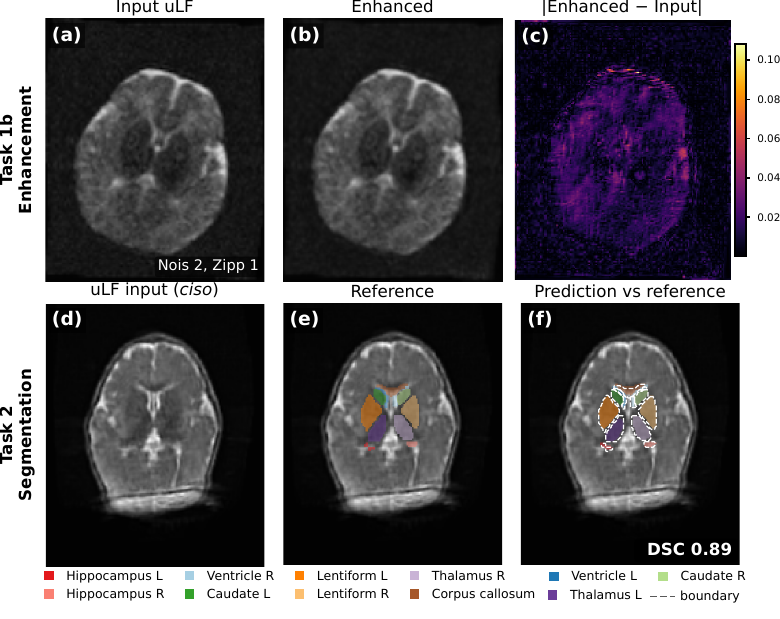}
\caption{Qualitative examples. Top (Task~1b): input, enhanced output, and residual map for a Noise$=$2, Zipper$=$1 case. Bottom (Task~2): input, reference, and predicted overlay (dashed reference boundary).}
\label{fig:qualitative}
\end{figure}

We report internal validation results across all three challenge tasks. Table~\ref{tab:results} summarizes the performance of the final integrated model that performs all tasks, while Table~\ref{tab:ablation} presents a compact ablation of the main architectural and training components. Figure~\ref{fig:qualitative} shows qualitative examples for Task~1b and Task~2.
The ablation demonstrates the benefit of the proposed coupled framework: compared with the baselines that handle only one task (segmentation or reconstruction alone) and the detached baseline, the full model improves 9 of the 10 metrics in Table~\ref{tab:ablation}, with DSC increasing monotonically across variants (.763 $\rightarrow$ .824 $\rightarrow$ .881). The BRISQUE$_\Delta$ exception reflects a tradeoff between perception and distortion: the strongest denoiser is also the smoothest, which BRISQUE penalizes more than PSNR, LPIPS, FID, or CLIPIQA do. Removing a component degrades the quality head the same way: dropping the residual between raw and reconstructed inputs and using a simpler QC head decreases the macro-F1 score from 0.64 to 0.54 and the macro recall from 0.63 to 0.55.
Finally, the official LISA validation leaderboard shows only small performance differences among the top methods across all three tasks, highlighting the challenge's competitiveness. We therefore emphasize the proposed framework's consistent performance across enhancement, quality assessment, and segmentation, aside from the explainable BRISQUE$_\Delta$ exception noted above.

\section{Discussion}
Decoupling reconstruction and segmentation into a regulated dual stream lets one network serve all three LISA tracks without a single objective dominating shared features, and label propagation driven by registration is what makes segmentation feasible under sparse annotation. Conditioning each artifact prediction on detached parent logits is more expressive than independent classifiers, and the auxiliary class for brain foreground outside the target structures limits label bleeding by explicitly separating target structures, remaining brain tissue, and extracranial background.

Several limitations remain. The training pairs augmented from atlases and the anatomical consistency term both depend on a registration network applied to noisy uLF data, though atlases matched in geometry and histogram matching mitigate this; the anatomical consistency term itself also penalizes how smooth the deformation is ($\|\nabla\phi\|_2^2$) rather than its magnitude directly, so adding an explicit penalty on magnitude is a natural extension. The artifact dependency graph is derived from cohort-level correlations and may not transfer to unseen sites, and strong label imbalance makes F1/F2 sensitive to weighting and threshold choices, so per-attribute reporting matters (Table~\ref{tab:results}). Finally, given the number of proposed components, Table~\ref{tab:ablation} spans broad configurations rather than a full leave-one-out study, and omits comparison against single-task baselines; both are left for future work.

\section{Conclusion}
We presented a unified 3D framework for artifact quality assessment, image enhancement, and segmentation of multiple subcortical structures in $0.064\,\mathrm{T}$ pediatric MRI for the LISA~2026 Challenge. The framework combines reconstruction that is aware of frequency content, regulated coupling between reconstruction and segmentation, propagation of atlas labels driven by registration, an auxiliary class for brain foreground outside the target structures, and a quality head conditioned on a graph and trained on detached reconstruction residuals and segmentation features. The deployed model performs reconstruction, segmentation, and quality assessment in a single inference pipeline; the registration network is used only during training, for supervision based on atlases and for anatomical regularization. Internal validation and ablation results support the feasibility of the integrated design.

\begin{credits}

\subsubsection{\discintname}
The authors have no competing interests to declare that are relevant to the content of this article.
\end{credits}

\bibliographystyle{splncs04}
\bibliography{ref}

\end{document}